\documentclass[sigconf]{acmart}
\AtBeginDocument{%
  }

\copyrightyear{2026}
\acmYear{2026}
\setcopyright{cc}
\setcctype{by}
\acmConference[UAVM '26]{The 4th Workshop on UAVs in Multimedia: Capturing the World from a New Perspective}{November 10--14, 2026}{Rio de Janeiro, Brazil}
\acmBooktitle{The 4th Workshop on UAVs in Multimedia: Capturing the World from a New Perspective (UAVM '26), November 10--14, 2026, Rio de Janeiro, Brazil}
\acmDOI{10.1145/3841459.3841550}
\acmISBN{979-8-4007-2948-5/2026/11}

\begin{document}

\title{AirAlign: Geometry-Aware Relative Pose Alignment for UAV Last-Meter Navigation}

\author{Jinyi Zhou}
\affiliation{%
  \institution{Nanjing University of Aeronautics and Astronautics}
  \city{Nanjing}
  \country{China}}
\email{jyzhou@nuaa.edu.cn}

\author{Shuo Feng}
\affiliation{%
  \institution{Nanjing University of Aeronautics and Astronautics}
  \city{Nanjing}
  \country{China}}
\email{fengshuo@nuaa.edu.cn}

\author{Yufei Wu}
\affiliation{%
  \institution{Nanjing University of Aeronautics and Astronautics}
  \city{Nanjing}
  \country{China}}
\email{wuyufei@nuaa.edu.cn}

\author{Piji Li}
\authornote{Corresponding author.}
\affiliation{%
  \institution{Nanjing University of Aeronautics and Astronautics}
  \city{Nanjing}
  \country{China}}
\email{pjli@nuaa.edu.cn}

\renewcommand{\shortauthors}{Jinyi Zhou, Shuo Feng, Yufei Wu, and Piji Li}

\begin{abstract}
Unmanned aerial vehicle (UAV) navigation in modern low-altitude environments requires more accurate pose alignment in the final approach stage for target information acquisition or manipulation, making "last-meter" navigation increasingly important. However, severe viewpoint and appearance variations make this task challenging. To tackle this problem, we propose AirAlign, a framework for RGB-only image-pair relative pose alignment for UAVs. AirAlign uses a pretrained visual geometry reconstruction model as the backbone to extract geometry-aware features from source-target image pairs. In addition, to better utilize the limited training data, we split the training set into multiple scene-disjoint folds for unseen cross-validation and model selection. During inference, the predictions of the selected models are averaged to form the ensemble output of the overall framework. Experiments on the PairUAV challenge at the ACMMM 2026 Workshop on UAVs in Multimedia demonstrate the effectiveness and robustness of our method, while comprehensive ablation studies validate the contribution of each component.
\end{abstract}

\begin{CCSXML}
<ccs2012>
<concept>
<concept_id>10010147.10010178.10010224.10010240.10010241</concept_id>
<concept_desc>Computing methodologies~Image representations</concept_desc>
<concept_significance>500</concept_significance>
</concept>
<concept>
<concept_id>10010147.10010178.10010224.10010245.10010254</concept_id>
<concept_desc>Computing methodologies~Reconstruction</concept_desc>
<concept_significance>500</concept_significance>
</concept>
<concept>
<concept_id>10010147.10010178.10010224.10010225.10010233</concept_id>
<concept_desc>Computing methodologies~Vision for robotics</concept_desc>
<concept_significance>500</concept_significance>
</concept>
</ccs2012>
\end{CCSXML}

\ccsdesc[500]{Computing methodologies~Vision for robotics}
\ccsdesc[500]{Computing methodologies~Image representations}
\ccsdesc[500]{Computing methodologies~Reconstruction}

\keywords{UAV last-meter navigation, visual geometry reconstruction}

\maketitle

\section{Introduction}
Unmanned aerial vehicles (UAVs) are increasingly deployed in low-altitude applications such as infrastructure inspection, search and rescue, and logistics delivery \cite{hassanalian2017classifications, shakhatreh2019unmanned}. Different from conventional navigation tasks, these scenarios require UAVs not only to move close to the target location, but also to achieve more accurate pose alignment during the final approach phase to support potential interaction with the target. This task, which is often referred to as "last-meter" navigation \cite{deuser2026UVA}, is important for practical UAV applications, since small errors in position or heading may affect landing safety, delivery accuracy, or the quality of close-range observation.

The challenge of this task lies in the fact that GNSS, which is commonly used in navigation, is usually insufficient for fine-grained pose estimation, especially when the UAV approaches the target at close range. Moreover, the source image and target image are captured from different altitudes and headings. Even when they observe the same target region, the appearance of buildings and their spatial arrangement can change significantly. Such visual variations make it difficult to recover the spatial relationship between the two views from appearance textures or global semantic features. In addition, the prediction pipeline is expected to remain lightweight for autonomous UAV flight by relying only on RGB images without additional depth sensors or auxiliary modalities. This setting makes it difficult to explicitly capture 3D geometric relationships, further increasing the challenge of accurate relative pose alignment. 

Direct studies on UAV "last-meter" navigation remain limited. Existing works focus mainly on place-level cross-view localization and image retrieval \cite{zheng2020university, zhu2022transgeo, deuser2023sample4geo}, where the goal is to match images across drone, satellite, or street-view platforms. Previous UAVM challenge solutions also follow this retrieval-based formulation \cite{liu2025matching, zhang2025vici}. Although these works address visual matching under large viewpoint variations, locating the target alone is still inadequate to support accurate pose alignment in the final stage of aerial navigation, where geometric information plays a crucial role.

To address these limitations, we propose AirAlign, an image-pair relative pose estimation framework for RGB-only UAV "last-meter" navigation, where the task is formulated as predicting the relative translation and heading from a source-target image pair. Specifically, we use a pretrained visual geometry reconstruction model \cite{wang2025pi} as the backbone of our framework to utilize its geometry-aware representations. Furthermore, the training set is evenly split into multiple scene-disjoint folds for cross-validation. Based on these folds, we train multiple models independently and use the average of their predictions as the ensemble output of the framework.

In summary, our main contributions are as follows:

\begin{itemize}
    \item We propose AirAlign, an image-pair relative pose alignment framework based on a pretrained visual geometry reconstruction model for UAV "last-meter" navigation, enabling fine-grained pose alignment during the final approach phase.

    \item We split the original training set into multiple folds for unseen cross-validation to select the best model in each round, and use the selected models for ensemble prediction.

 	\item The strong performance on the PairUAV challenge test set demonstrates the effectiveness of our framework, with ablation studies validating the contribution of each component.
\end{itemize}

\section{Related Work}
\subsection{Visual Geometry Reconstruction}
Traditional Structure-from-Motion (SfM) \cite{cui2017hsfm, schonberger2016structure} and Multi-view Stereo (MVS) \cite{furukawa2015multi, schonberger2016pixelwise} pipelines recover 3D structure through feature matching and dense stereo reconstruction, while recent feed-forward visual geometry models \cite{wang2024dust3r, wang2025vggt} provide a more direct alternative by predicting geometric representations from images. Building on this line of work, MASt3R \cite{leroy2024grounding} further introduces correspondence modeling between image pairs, grounding image matching in 3D geometry rather than relying only on appearance similarity. Inspired by these works, we argue that spatial geometric information is also important for image-pair relative pose estimation and alignment. Therefore, we adopt the recent permutation-equivariant visual geometry reconstruction model \cite{wang2025pi} for this task.

\subsection{Aerial Object-Goal Navigation}
Goal-oriented aerial navigation requires UAVs to explore environments and locate the target. UAV-ON \cite{xiao2025uav} and CityNav \cite{lee2025citynav} use natural language instructions as goal descriptions, while OctMem-Agent \cite{zhou2026memory} further improves performance with memory-augmented exploration and localization. For image-goal navigation, ANWM \cite{zhang2025aerial} uses a world model to support navigation planning, and GeoExplorer \cite{mi2025geoexplorer} introduces a curiosity-driven mechanism to encourage active exploration. However, these works mainly focus on reaching the target vicinity rather than achieving fine-grained pose alignment. Therefore, our work addresses an image-pair relative pose estimation problem, where the model predicts relative translation and heading from a source-target UAV-view image pair.

\section{Methodology}
\subsection{Overall Architecture}
\begin{figure*}[t]
    \centering
    \includegraphics[width=\textwidth]{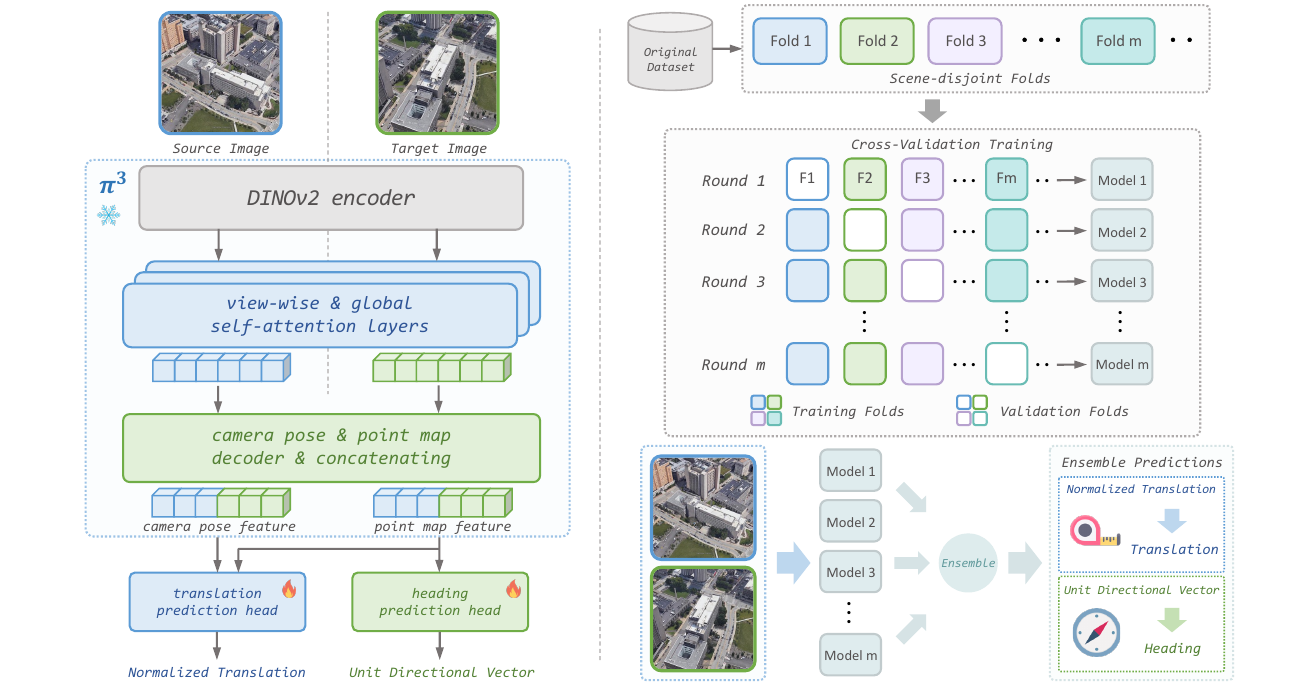}
    \caption{Overview of AirAlign. The left part shows the model architecture, while the right part illustrates the scene-disjoint cross-validation training strategy and ensemble inference process.}
    \label{fig:framework}
\end{figure*}

We argue that predicting relative translation and heading from an RGB-only image pair requires strong spatial geometric understanding, rather than relying only on simple keypoint matching. In our framework, a pretrained visual geometry reconstruction model \cite{wang2025pi} is used as the backbone to extract features related to camera poses and 3D point maps from the source-target RGB image pair, with details provided in Section \ref{sec:feature}. Both types of features are then fed into the translation prediction head to provide distance information between the camera and the observed target, while the camera pose features are used separately for simpler heading prediction, which is described in Section \ref{sec:prediction}. We further evenly split the original training set into multiple scene-disjoint folds for cross-validation, and use the trained models jointly for prediction, as discussed in Section~\ref{sec:training}. The overall framework is illustrated in Figure~\ref{fig:framework}.

\subsection{Geometry-Aware Feature Extraction}
\label{sec:feature}
We use a pretrained visual geometry reconstruction model for geometry-aware feature extraction. Specifically, we adopt the main components of $\pi^3$ \cite{wang2025pi} as the backbone, where the DINOv2 \cite{oquab2023dinov2} encoder and a series of alternating view-wise and global self-attention layers form the geometry feature extractor, and two dedicated decoders are used to produce camera pose and 3D point map representations, respectively. These components are used to obtain features of the relative spatial relationships from the input image pair.

Formally, given a source image $I_s$ and a target image $I_t$, the backbone first maps the image pair into hidden geometry features:
\begin{equation}
    \mathbf{H}_s, \mathbf{H}_t = F_{\pi^3}(I_s, I_t),
\end{equation}
where $F_{\pi^3}$ denotes the geometry feature extractor. Then, the camera pose and point map decoders produce the corresponding features:
\begin{equation}
    \mathbf{c}_s, \mathbf{c}_t = D_{cam}(\mathbf{H}_s, \mathbf{H}_t), \quad
    \mathbf{p}_s, \mathbf{p}_t = D_{pts}(\mathbf{H}_s, \mathbf{H}_t),
\end{equation}
where $D_{cam}$ and $D_{pts}$ denote the camera pose decoder and the point map decoder, respectively. The pair-level geometry-aware representations are then formed by concatenating the source and target image features:
\begin{equation}
    \mathbf{f}_{cam} = [\mathbf{c}_s; \mathbf{c}_t], \quad
    \mathbf{f}_{pts} = [\mathbf{p}_s; \mathbf{p}_t].
\end{equation}
Here, $[\cdot;\cdot]$ denotes concatenation along the feature dimension.

\subsection{Relative Translation and Heading Prediction}
\label{sec:prediction}
For the task of relative translation and heading prediction, we replace the original output heads of $\pi^3$ with two task-specific heads. Both heads are implemented as three-layer MLPs with two hidden layers. The heading prediction head takes only the camera pose feature $\mathbf{f}_{cam}$ as input, and its output is designed as a two-dimensional unit direction vector of the relative angle, which reduces the training complexity caused by directly regressing the angle. This process is formulated as:
\begin{equation}
    \hat{\mathbf{h}} = \frac{\mathrm{MLP}_{head}(\mathbf{f}_{cam})}{\|\mathrm{MLP}_{head}(\mathbf{f}_{cam})\|_2}.
\end{equation}

The translation prediction head takes both $\mathbf{f}_{cam}$ and $\mathbf{f}_{pts}$ as input to obtain a more comprehensive geometric representation:
\begin{equation}
    \hat{t} = \sigma\left(\mathrm{MLP}_{trans}([\mathbf{f}_{cam}; \mathbf{f}_{pts}])\right),
\end{equation}
where $\sigma(\cdot)$ is the sigmoid function, which maps the output to a normalized range.

\subsection{Training Strategy and Model Ensemble}
\label{sec:training}
To fully utilize the training data and improve the robustness of the framework, we evenly split the original training set provided by the challenge into multiple folds, where the scenes in different folds do not overlap. Each fold is used in turn as an unseen validation set to select the model checkpoint with the lowest final score, as defined in Section~\ref{sec:metrics}, across all training epochs, while the remaining folds are used as the training set to train the model for the current round. During training, the pretrained $\pi^3$ backbone is frozen, and only the prediction heads are optimized.

For the heading prediction head, we use smooth L1 loss on the normalized two-dimensional direction vectors:
\begin{equation}
    \mathcal{L}_{head}
    =
    \operatorname{smooth}_{L_1}(\hat{\mathbf{h}} - \mathbf{h}),
\end{equation}
where $\hat{\mathbf{h}}$ denotes the predicted unit direction vector of the relative heading, and $\mathbf{h}$ denotes the ground-truth direction vector:
\begin{equation}
    \hat{\mathbf{h}} = (\cos\hat{\theta}, \sin\hat{\theta}),
    \quad
    \mathbf{h} = (\cos\theta, \sin\theta),
\end{equation}
where $\hat{\theta}$ and $\theta$ are the predicted and ground-truth relative headings, respectively. We further introduce the relative angle error loss as an auxiliary term:
{\small
\begin{equation}
    \mathcal{L}_{\mathrm{rel\text{-}ang}}
    =
    \frac{1}{N}\sum_{i=1}^{N}
    \frac{
    \min\left(
    |\hat{\theta}_{i,\mathrm{mod}} - \theta_{i,\mathrm{mod}}|,
    360 - |\hat{\theta}_{i,\mathrm{mod}} - \theta_{i,\mathrm{mod}}|
    \right)
    }
    {|\theta_{i,\mathrm{mod}}| + \epsilon},
\end{equation}
}
where both angles are normalized as
\begin{equation}
    \hat{\theta}_{i,\mathrm{mod}}
    =
    \operatorname{mod}(\hat{\theta}_i, 360),
    \quad
    \theta_{i,\mathrm{mod}}
    =
    \operatorname{mod}(\theta_i, 360).
\end{equation}

The loss for the translation prediction head also consists of two terms. The first term is the smooth L1 loss on the normalized translation prediction:
\begin{equation}
    \mathcal{L}_{trans}
    =
    \operatorname{smooth}_{L_1}(\hat{t} - \bar{t}),
\end{equation}
where $\hat{t}$ and $\bar{t}$ are the predicted and ground-truth normalized translations. The second term is the relative distance error:
\begin{equation}
    \mathcal{L}_{\mathrm{rel\text{-}dist}}
    =
    \frac{1}{N}\sum_{i=1}^{N}
    \frac{|\tilde{t}_i - t_i|}{|t_i| + \epsilon},
\end{equation}
where $t_i$ is the original ground-truth translation and $\tilde{t}_i$ is the denormalized prediction:
\begin{equation}
    \tilde{t}_i = \hat{t}_i(t_{\max} - t_{\min}) + t_{\min}.
\end{equation}
$\bar{t}_i$ is obtained by normalizing $t_i$ with the same range parameters.

The overall training loss is formulated as:
\begin{equation}
    \mathcal{L}
    =
    \mathcal{L}_{head}
    + \lambda_{\mathrm{ang}}\mathcal{L}_{\mathrm{rel\text{-}ang}}
    + \mathcal{L}_{trans}
    + \lambda_{\mathrm{dist}}\mathcal{L}_{\mathrm{rel\text{-}dist}},
\end{equation}
where $\lambda_{\mathrm{ang}}$ and $\lambda_{\mathrm{dist}}$ are the weights of the relative angle error loss and the relative distance error loss, respectively.

For inference, the final output of the framework is constructed by averaging the predictions from multiple trained models. The ensemble heading prediction is obtained by averaging the predicted two-dimensional direction vectors and recovering the final angle:
\begin{equation}
    \mathbf{h}_{ens}
    =
    \frac{1}{M}
    \sum_{m=1}^{M}
    \hat{\mathbf{h}}^{(m)},
    \quad
    \theta_{ens}
    =
    \operatorname{atan2}(h_{ens,2}, h_{ens,1}).
\end{equation}
The ensemble translation prediction is computed as:
\begin{equation}
    t_{ens}
    =
    \frac{1}{M}
    \sum_{m=1}^{M}
    \tilde{t}^{(m)}.
\end{equation}
Here, $M$ denotes the number of models in the ensemble.

\section{Experiments}
\subsection{Implementation Details}
Our model is trained on the PairUAV \cite{deuser2026UVA} training set and takes images with a resolution of $448 \times 448$ as input. The $\pi^3$ backbone uses frozen pretrained weights from \cite{wang2025pi}, while the task heads are randomly initialized and trained. To improve training efficiency, we first use the backbone to extract geometry-aware features and cache them to avoid repeated forward passes. During training, these cached features are directly used as inputs to the task heads. The batch size is set to 2048, and the task heads are optimized for 550 epochs in each training round using AdamW with an initial learning rate of $2 \times 10^{-3}$. Following the official baseline \cite{li2026lastmeterprecisionnavigationuavs}, we set $t_{\min}$ to $-132$ and $t_{\max}$ to $132$. $\lambda_{\mathrm{ang}}$ and $\lambda_{\mathrm{dist}}$ are set to 0.4 and 0.6.

\subsection{Evaluation Metrics}
\label{sec:metrics}
The challenge evaluation metrics consist of the distance relative error and the angle relative error. For each valid sample, the distance relative error is computed as:
\begin{equation}
    e^{dist}
    =
    \frac{|\hat{t} - t|}{|t|},
\end{equation}
where $\hat{t}$ and $t$ denote the predicted and ground-truth translations.

For the heading angle, both the predicted and ground-truth angles are first normalized to $[0, 360)$ as $\hat{\theta}_\mathrm{mod}$ and $\theta_\mathrm{mod}$. The angle relative error is then computed using the minimum circular difference:
{\small
\begin{equation}
    e^{ang}
    =
    \frac{
    \min\left(
    |\hat{\theta}_\mathrm{mod} - \theta_\mathrm{mod}|,
    360 - |\hat{\theta}_\mathrm{mod} - \theta_\mathrm{mod}|
    \right)
    }
    {|\theta_\mathrm{mod}|}.
\end{equation}
}

The final score for each valid sample is the average of the relative errors:
\begin{equation}
    s = \frac{e^{dist} + e^{ang}}{2}.
\end{equation}

The leaderboard score averages the final scores over all valid samples.

\subsection{Overall Performance}
In the PairUAV challenge at the ACMMM 2026 Workshop on UAVs in Multimedia, AirAlign achieves a final score of 0.002790, with a distance relative error of 0.003175 and an angle relative error of 0.002405. As shown in Table~\ref{tab:main_results}, our final result significantly outperforms the official baseline of 0.633957 and ranks 6th on the final leaderboard, demonstrating the strong capability and robustness of our framework for UAV image-pair relative pose alignment.

\begin{table}[H]
    \centering
    \caption{Evaluation results on the challenge test set.}
    \label{tab:main_results}
    \setlength{\tabcolsep}{7pt}
    \begin{tabular}{lcccc}
        \toprule
        Team & Dist. Err. & Ang. Err. & Score & Rank \\
        \midrule
        light & 0.002277 & 0.000780 & 0.001529 & 1 \\
        liang & 0.003086 & 0.000378 & 0.001732 & 2 \\		
        ze\_rong & 0.001330 & 0.002419 & 0.001874 & 3 \\		
        g2502795c & 0.002222 & 0.002321 & 0.002272 & 4 \\		
        Team & 0.002186 & 0.002609 & 0.002398 & 5 \\		
        ginnne (ours) & 0.003175 & 0.002405 & 0.002790 & 6 \\		
        nuaaceie & 0.003029 & 0.003350 & 0.003189 & 7 \\		
        chenzhang & 0.004594 & 0.002606 & 0.003600 & 8 \\		
        xinyan & 0.006237 & 0.001810 & 0.004023 & 9 \\
        Baseline & 0.988067 & 0.279847 & 0.633957 & -- \\
        \bottomrule
    \end{tabular}
\end{table}

\subsection{Ablation Study}
We evaluate the effects of the number of split folds, the auxiliary relative losses, and the design of the input features for the task heads on the overall framework. The results are shown in Table~\ref{tab:ablation}.

\paragraph{Effect of the number of split folds.}
We use 3, 5, and 7 split folds. The 5-fold setting used in our main method achieves the best final score of 0.002790, outperforming the 3-fold and 7-fold settings with scores of 0.004239 and 0.003883, respectively. This indicates that more folds do not necessarily improve performance. With 3 folds, fewer models and less training data per round may lead to insufficient training and less stable ensemble prediction, whereas fewer validation scenes under 7 folds may weaken model selection.

\paragraph{Effect of the auxiliary relative losses.}
The ablation results show that both auxiliary relative losses contribute to the final performance. After removing the corresponding relative losses, the distance error and the angle error increase, respectively, by 0.008840 and 0.001196, indicating that the design of the auxiliary losses is important for improving the accuracy of the relative pose alignment.

\paragraph{Effect of the input features.}
The input feature ablations show that different tasks require different geometric features. Adding point map features to the heading head increases the angle error to 0.003691, indicating that heading prediction mainly relies on camera pose features, while point map features may introduce redundancy or interference. For translation prediction, removing either feature degrades performance and removing camera pose features causes a larger increase in distance error to 0.011359. Thus, translation prediction benefits from both camera pose and point map features.

\begin{table}[t]
    \centering
    \caption{Ablation studies of the proposed framework.}
    \label{tab:ablation}
    \setlength{\tabcolsep}{4pt}
    \begin{tabular}{lccc}
        \toprule
        Setting & Dist. Err. & Ang. Err. & Score \\
        \midrule
        \textbf{AirAlign} & \textbf{0.003175} & \textbf{0.002405} & \textbf{0.002790} \\
        \midrule
        \multicolumn{4}{l}{\textit{Effect of the number of split folds}} \\
        3-fold ensemble & 0.004853 & 0.003625 & 0.004239 \\
        5-fold ensemble (main) & 0.003175 & 0.002405 & 0.002790\\
        7-fold ensemble & 0.004126 & 0.003641 & 0.003883 \\
        \midrule
        \multicolumn{4}{l}{\textit{Effect of the auxiliary relative losses}} \\
        w/o relative angle loss & 0.003324 & 0.003601 & 0.003463 \\
        w/o relative distance loss & 0.012015 & 0.002489 & 0.007252 \\
        \midrule
        \multicolumn{4}{l}{\textit{Effect of the input features}} \\
        Head. head with point feat. & 0.004201 & 0.003691 & 0.003946 \\
        Trans. head w/o camera feat. & 0.011359 & 0.002440 & 0.006900 \\
        Trans. head w/o point feat. & 0.007748 & 0.002432 & 0.005090 \\
        \bottomrule
    \end{tabular}
\end{table}

\section{Conclusion}
We propose AirAlign, a relative pose alignment framework for UAV "last-meter" navigation based on a pretrained visual geometry reconstruction model. We split the training set into scene-disjoint folds for model training and selection, and average the predictions from the selected models during inference. AirAlign achieves a high ranking in the PairUAV challenge at the ACMMM 2026 Workshop on UAVs in Multimedia, demonstrating its effectiveness.
\balance
\bibliographystyle{ACM-Reference-Format}
\bibliography{references}

\end{document}